\pdfoutput=1
\documentclass[11pt,letterpaper]{article}
\usepackage{emnlp2016}
\usepackage{times}
\usepackage{latexsym}
\usepackage[utf8]{inputenc}
\usepackage[T1]{fontenc}
\usepackage{tablefootnote}
\usepackage{gb4e}

\let\tempone\itemize
\let\temptwo\enditemize
\renewenvironment{itemize}{\tempone\addtolength{\itemsep}{-0.7\baselineskip}}{\temptwo}

\let\tempone\xlist
\let\temptwo\endxlist
\renewenvironment{xlist}{\tempone\addtolength{\itemsep}{-1.4\baselineskip}}{\temptwo}

\newcommand{\enotesoff}{\long\gdef\enote##1##2{}}

\enotesoff
\emnlpfinalcopy

\def\emnlppaperid{***}

\title{Challenges of Computational Processing of Code-Switching}

 \author{Özlem Çetinoğlu \and Sarah Schulz \and Ngoc Thang Vu \\
         IMS, University of Stuttgart\\Germany\\\tt{\{ozlem,schulzsh,thangvu\}@ims.uni-stuttgart.de}}
         
\date{}

\begin{document}

\maketitle

\begin{abstract}
This paper addresses challenges of Natural Language Processing (NLP) on non-canonical multilingual data in which two or more languages are mixed.
It refers to \textit{code-switching} which has become more popular in our daily life and therefore obtains an increasing amount of attention from the research community. 
We report our experience that covers not only core NLP tasks such as normalisation, language identification, language modelling, part-of-speech tagging and dependency parsing but also more downstream ones such as machine translation and automatic speech recognition.
We highlight and discuss the key problems for each of the tasks with supporting examples from different language pairs and relevant previous work.
\end{abstract}

\section{Introduction}
\label{sec:intro}

Data that includes mixing of two or more languages finds more place 
in the Natural Language Processing (NLP) tasks over the last few
years. This changing picture induces its own challenges as well.

The analysis of mixed language is not a new field, and has been extensively studied 
from several sociological and linguistic 
aspects \cite{Poplack:1980,MyerScotton:1993,Muysken:2000,auer2007,bullock2012}. 
This has also brought different perspectives
on the definition and types of mixed language. Switching between 
sentences (\textit{inter-sentential}) is distinguished from
switching inside of one sentence (\textit{intra-sentential}). 
\newcite{Poplack:1980} defines 
\textit{code-switching} as `the alternation of two languages within a 
single discourse, sentence or constituent'. \newcite{Muysken:2000} avoids this 
term arguing that it suggests alternation but not insertion, and prefers 
\textit{code-mixing} for intra-sentential switching. \newcite{MyerScotton:1993}
employs the cover term \textit{code-switching} for the use of two languages in the 
same conversation, sentence, or phrase. In this paper we use code-switching (CS) 
as a cover term for all types of mixing. The terminology is still controversial
among researchers, but there is no doubt that all types pose challenges 
for computational systems built with monolingual data.

Computational approaches in the analysis of CS data are quite recent as
compared to linguistic studies. The first theoretical framework to parse code-switched 
sentences dates back to the early 1980s \cite{Joshi:1982}, yet few studies are 
done in the 2000s \cite{goyal:2003,Sinha2005,Solorio:2008,Solorio:2008b}.
With the beginning of the last decade, this picture has changed due to
 increasingly multi-cultural societies and the rise of social media. Supported 
with the introduction of annotated data sets on several language pairs, different
tasks are applied on CS data. 

The characteristics of mixed data affect tasks in different ways,
sometimes changing the definition (e.g. in language identification, the shift from
document-level to word-level), sometimes by creating new lexical and syntactic 
structures (e.g. mixed words that consist of morphemes from two different languages).
Thus, it is clear that mixed data calls for 
dedicated tools tailored to the specific problems and contexts encountered. In order to take these specialties
into account, these different cases have to be understood. This way, differences in techniques for 
monolingual and mixed language processing can be unfolded to 
yield good results. 

In this paper, we view CS processing from a variety of perspectives, and discuss the unique
challenges that one encounters. We redescribe NLP tasks under the assumption that the 
data contains more than one language. For tasks that are studied more compared to 
others we compile approaches taken by previous work. Examples from different language pairs 
highlight the challenges, supporting the need for awareness about the nature of mixed data for 
successful automatic processing.





\section{Data}
\label{sec:data}
\noindent
\textbf{Nature of the data} Annotated CS corpora, that are designed for computational purposes, center around three sources so far: spoken data \cite{Solorio:2008b,Lyu2015,yilmaz:2016}, 
social media \cite{nguyen:2013,barman:2014,vyas:2014,Solorio2014,fire:2014,jamatia:2015,cetinoglu:2016a,samih:lrec16}, 
and historical text \cite{Schulz:2016b}. 
All these data sources are challenging on their own even if 
they do not exhibit CS. They are non-canonical in their orthography, lexicon, and 
syntax, thus the existing resources and tools should be adapted, or new ones
should be created to handle domain-specific issues in addition to the 
challenges of CS itself.

\enote{ozlem}{are there any other resources to add to the list?}


\noindent
\textbf{Accessing the data} Although CS is prominent in every day life, especially in countries
with a high percentage of multilingual communities, accessing it is still 
problematic. Speech as one of the main sources requires consent prior
to recording. One way to keep recordings as natural as possible is to not mention
the goal as capturing CS instances to participants. Being recorded however raises self-awareness,
and could possibly change how the language is used. Many bilinguals are not keen 
on mixing languages, e.g. human annotators comment ``we shouldn’t code-switch'' \cite{Solorio:2008}. 

On this point, social media has an advantage: users of Facebook, Twitter, forums, or blogs, 
are not aware that their data will be used for analysis, which therefore makes it a more 
naturalistic setting. They give their consent 
after, once the content is created. Among social media sites, Twitter has its 
disadvantages like license issues, and limited characters per tweet. Other media that
does not have these advantages remain popular sources.

\section{Normalisation}
\label{sec:normalisation}

Text normalisation is the task of standardising text that deviates from some 
agreed-upon (or canonical) form. This can e.g. refer to normalising social media language to 
standard language (``newspaper language'') (cf. e.g. \newcite{Schulz:2016} and \newcite{Aw2006})
or historical language to its modern form \cite{Bollmann2011}.
Since mixed text often occurs in spoken language or text close in nature to spoken language like 
social media, normalisation is a highly relevant task for the processing of such text.
In the case of mixed text there are two languages embedded into each other. 
Defining a canonical form is a challenge because each of the languages 
should be standardised to its own normal form.

Normalisation of text has started out as a task mainly solved on token-level.
Most of the recent approaches are based on context e.g. in character-based 
machine translation \cite{Liu:2012,Schulz:2016}. This results
from the fact that normalisation requires a certain degree of semantic disambiguation of words in context
to determine if there is a normalisation problem. These problems can appear
on two levels: a) The word is an out-of-vocabulary (OOV) word (which are the easy cases), thus it does not exist 
in the lexicon. b) The word is the wrong word in context (often just the 
wrong graphematic realisation e.g. \textit{tree} instead of \textit{three}).

This context dependency results in issues for mixed text.
The presence of CS increases the number of possible actions regarding an 
erroneous word. The word could be incorrect in one language and not in the other, or
incorrect in both. Either way, the intended language should be decided as well as the usage. 
(\ref{ex:meisten}) emphasises the need for semantic understanding in context \cite{cetinoglu:2016a} 
in which Turkish (in italics) is mixed with German (in bold).

\begin{exe}
\begin{small}
\ex\label{ex:meisten} 
 \gll \textit{\textbf{meisten}} \textit{kıyımıza}  \textit{vurmuş}  \textit{olmasi}  \textit{muhtemel} :) \\
 Meis.Abl(TR)/mostly(DE) shore.P1pl.Dat hit.EvidPast be.Inf possible\\
 \trans `It is possible that it hit our shore from Meis.'/`Mostly it is possible that it hit our shore.'  
\end{small} 
\end{exe}

The first word in the example can be interpreted in two different ways. In Turkish
it could be an orthographically incorrect form of \textit{Meis'ten} `from Meis' which refers to
the Greek island Kastellorizo. Or it could be a typo where the \textit{s} of
German \textit{meistens} `mostly' is missing. Such uncertainty might be observed more
when language pairs are from similar languages, and share identical and similar words.

Another example is taken from a corpus of Facebook chats \cite{androutsopoulos:2015}. 
In this example, three languages, Greek (in italics), German (in bold) and English, 
are used within one sentence:
\begin{exe}
\begin{small}
\ex\label{ex:greek} 
 \gll hahahahaha \textit{\textbf{ade}}      ok \textbf{tanz} \textit{zebekiko} \textbf{aber} \textbf{bei} billy jean please \\
      hahahahaha {come on(GR)/goodbye(DE)} okay(GR/DE/EN) dance zebekiko but on Billy Jean please\\
 \trans `hahahahaha come on ok dance zebekiko but on Billy Jean please'  
\end{small}
\end{exe}

\newcite{androutsopoulos:2015} explains that the post starts with a bilingual 
discourse marker that indexes concessiveness (\textit{ade ok} ‘come on, ok’). The Greek 
vernacular item \textit{ade} is combined with \textit{ok}, which could be assigned to any 
of the three languages whereas the preceding \textit{hahahahaha} is not a word 
in any of them. \textit{Ade} `goodbye', however, is an existing German word 
and without larger context  of the sentence it is hard 
to determine if the German \textit{Ade} is intended, in which case a 
normalisation action (capitalisation) is required, or indeed the Greek 
vernacular \textit{ade}. Semantic contextualisation is aggravated due 
to the trilingual context. 

One solution is the approach by \newcite{Zhang2014}. They use a neural 
net based translation architecture for Chinese-English text normalisation.
 It includes a Chinese language model and a
Chinese-English translation model as well as user history-based context.
Since training material for such systems might be sparse for some language pairs, methods for 
mixed text tend to return to smaller context windows as done by \newcite{Dutta2015}. 
They suggest to use two monolingual language models with context windows 
depending on the neighbouring words using language identification information. 
In case of a high density of switch points between languages, the context window might be small.

As another normalisation challenge, \newcite{Kaur:2015} describe issues emerging from mixing different scripts in 
Punjabi-English and \newcite{Sarkar:2016} for Hindi-English
and Bengali-English social media text. Since text is often realised in Roman script,
in order to utilise resources from other writing systems, the text has to be 
mapped back to the system of the respective language. Due to this problem
\newcite{barman:2014} do not use existing Bengali and Hindi resources in their
dictionary-based approach. \newcite{das:2014} Romanised the resources whereas
\newcite{vyas:2014} go in the opposite direction and develop a back-transliteration
component.

\section{Language Modelling}
\label{sec:lm}
A statistical language model assigns a probability to a given sentence or utterance. 
Models such as n-gram models \cite{Brown1992}, factored language models \cite{Bilmes2003}  and neural language models \cite{Bengio2003}  are used 
in many NLP applications such as machine translation and automatic speech recognition.
One can consider mixed text as an individual language and use existing techniques to train the model.
Tokenisation and normalisation are the first steps to prepare the training  data. 
Hence, one will face the problems presented in Section \ref{sec:normalisation} first.
Another serious problem is the lack of CS training data which makes statistical language models unreliable.
We consider three kinds of CS to identify challenges of code-switching language modelling (LM).\\
\noindent
\textbf{Inter-sentential CS} In addition to the CS data corpus, we can use monolingual data to train several language models and interpolate them.
Under the assumption that we can find monolingual data which has the same domain as the CS data, there is no obvious problem in this case.\\
\noindent
\textbf{Intra-sentential CS} The only available data resource is the code-switching training data.
Previous work suggests to add syntactic and semantic information into the original statistical model to first predict the CS points and then the current word.
It shows improvement when integrating additional resources such as language identification and POS tags into the recurrent neural language model \cite{AdelVu2013}, the factored language model \cite{Adel2015} and their combination \cite{Adel2013}.
Their statistical analysis in Table \ref{triggerPos} on the Mandarin-English CS data set \cite{Lyu2015} gives some insights on how accurate one can predict CS points based on POS tags.
\begin{table}[h!]
\begin{center}
\begin{small}
\begin{tabular}{llrr}
Tag & Meaning & Frequency & CS-rate\\
\hline \hline
DT   & determiner &   11276 &  40.44\%\\
MSP  & other particle & 507 & 32.74\% \\
\hline
NN   & noun &   49060 &  49.07\% \\
NNS  & noun (plural)  &  4613  &  40.82\%  \\
RP   & particle & 330 & 36.06\% \\
RB   & adverb  &  21096 &  31.84\% 
\end{tabular}
\caption{\label{triggerPos} {Mandarin and English POS tags that trigger a code-switching (First two columns: Man $\rightarrow$ En, rest: En $\rightarrow$ Man)}}
\end{small}
\end{center}
\vspace{-0.3cm}
\end{table}
Additional information such as language identification and POS tags might not be accurate due to problems presented in Section \ref{sec:langid} and Section \ref{sec:postag}.
In their work, they propose to combine Stanford Mandarin and English POS taggers to generate POS tags.
There is, however, no report on POS tagger performance due to the lack of gold data. 

Li and Fung (2012;2014) \nocite{Li2012,Li2014} propose another research direction which assembles syntactic inversion constraints into the language model. 
Instead of learning to predict the CS points alone, they suggest to learn the permission probabilities of CS points from a parallel corpus as to not violate the grammatical rules of both languages.
This information is then integrated in a language model to constrain the CS points.
It appears to be a promising approach if a large amount of parallel data of the 
two language exists and if the assumption holds that people do not change the 
grammatical rules of the mixed languages.\footnote{This  assumption 
is quite controversial among CS researchers, even Section \ref{sec:parsing} has counter-examples 
that show grammar changes. The assumption, however, could still be useful in statistical
systems if the majority of switches follow the rules.}\\
\noindent
\textbf{Intra-word CS} In addition to challenges presented in the previous paragraphs, one has to face the out-of-vocabulary problem when CS appears within a word.
This word has a high potential to be an unknown word. 
For example in the German-Turkish corpus \cite{cetinoglu:2016a}, 1.16\% of the corpus are mixed words.
93.4\% of them appear only once which indicates a big challenge not only for language modelling but also for other tasks.

\section{Language Identification}
\label{sec:langid}
Identifying the language of a text as one of the given languages is 
considered to be a solved task \cite{McNamee2005}. 
Simple n-gram approaches \cite{cavnar:1994}, character encoding detection \cite{Dunning1994}  
or stop word lists \cite{grefenstette:1995} can lead to a recognition accuracy 
of up to 100\% on benchmark data sets. 

Discriminating between closely related languages that show a significant lexical and 
structural overlap, like Croatian and Serbian, already 
poses a bigger problem. Stop word list approaches are problematic in such language
pairs. N-gram approaches show an accuracy of up to 87\% \cite{Tan2014}.

However, all these techniques rely on the assumption that the input text is 
encoded in exactly one language. As soon as different languages 
are mixed inside a text or further within a sentence, a more fine-grained 
detection is needed. CS reduces the minimum unit of detection to a token. 

Language identification (LID) is the most well-studied task among computational CS approaches:
there is relatively more annotated data; it is one of the preprocessing steps for
more complex tasks; and shared tasks \cite{Solorio2014,fire:2014} attract more research.

Language identifiers with good performance on monolingual input \cite{cavnar:1994,lui:2012} 
encounter accuracy loss due to shorter and/or unobserved context \cite{nguyen:2013,volk:2014}.
Thus researchers have chosen to build new tools tailored to CS, using
simple dictionary-based methods or machine learning techniques such
as Naive Bayes, CRF, and SVM (Lignos and Marcus, 2013; Nguyen and Do\u{g}ru\"{o}z, 2013; 
Voss et al., 2014; Das and Gamb\"{a}ck, 2014; Barman et al., 2014 
and cf. Solorio et al., 2014).\nocite{lignos:2013,nguyen:2013,voss:2014,das:2014,barman:2014,Solorio2014} 
While they outperform language identifiers trained on monolingual data, 
they reach accuracies in the mid-90s. Shared task results \cite{Solorio2014} report
even lower F-scores (80-85\%), for some language pairs (Modern Standard Arabic- Egyptian Arabic,
and surprise data sets for Nepalese-English, Spanish-English).

Some of the challenges CS poses are inherent to the languages involved, which 
then propagate to language annotation and identification. 
One of the language-specific challenges is language annotation when the mixed 
languages are closely related either linguistically or historically, e.g., 
Modern Standard Arabic and dialects \cite{elfardy:2012,samih:lrec16} and 
Frisian-Dutch \cite{yilmaz:2016}. In such 
cases it is hard to find a clear distinction between code-switching and borrowing, 
thus deciding the language ID of a particular token. For English-Hindi, \newcite{das:2014} give
the word `glass' as an example. The concept was borrowed during the British
colonisation in India, and Indian dictionaries contain the transliterated
version. Yet, annotators sometimes labelled it as English. 

The opposite is also observed. Both \newcite{vyas:2014} and \newcite{barman:2014} 
propose to label English borrowings in Hindi and Bengali as English. However,
\newcite{barman:2014} report that some annotators still annotated them as Hindi 
and Bengali. In the end almost 7\% of the unique tokens were labelled
in more than one language during annotation, which demonstrates that it is
challenging to decide language IDs even for humans.

Vague division between CS and borrowing partially affects the language pairs 
when one or both are influenced by another language, e.g. English in present day. 
For instance in the Turkish-German corpus \cite{cetinoglu:2016a}, the word \textit{leggings} was controversial 
among annotators, as some think it is English while others believe it is 
already integrated in the German language. This phenomenon could be challenging 
for statistical systems too, if the monolingual resources contain those 
controversial words inconsistently or in the opposite label of gold data.

Another big challenge for LID is mixing
two languages inside one single word. These mixed words are treated differently
among researchers. While many do not specify a special tag for intra-word
mixing due to very infrequent representation in their corpus, \newcite{das:2014} propose 
10 tags that mark the combinations of root and suffix languages. 
The CodeSwitch Workshop 2014 Shared Task \cite{maharjan-EtAl:2015:LAW}, \newcite{barman:2014}, and
\newcite{cetinoglu:2016a} use a Mixed tag. 

This pattern is e.g. very productive in German-Turkish code-switching where 
the suffixes of Turkish, as an agglutinating language, are attached to 
German words.\footnote{The Turkish-German CS tweets \cite{cetinoglu:2016a} have 1.16\%
Mixed tokens as compared to 0.32\% Mixed in En-Bn-Hi \cite{barman:2014} and 
0.08-0.60\% in Ne-En and 0.04-0.03\% in Es-En \cite{maharjan-EtAl:2015:LAW} corpora.}
This can result in words 
like \textit{Aufgabeler} `assignments' in (\ref{ex:aufgabe}) where the Turkish plural suffix \textit{-ler} 
is appended to the German word \textit{Aufgabe} `assignment' and poses problems
not only for LID but also for existing tools for POS tagging
and morphological analysis.


\section{POS Tagging}
\label{sec:postag}

POS tagging assigns a category from a given set to each input token. It has 
a popular use as a standalone application or as part of a preprocessing step for
other tasks, e.g., parsing. It is the second most popular task after language 
identification in the current state of CS research. Unlike LID,
CS does not change the definition of the task. Nevertheless,
the task gets harder compared to tagging monolingual text. While state-of-the-art models
reach over 97\% accuracy on canonical data\footnote{\texttt{https://aclweb.org/aclwiki/index.php?\\
title=POS\_Tagging\_(State\_of\_the\_art)}},
in work on CS data scores mostly around 70\% are reported.

One problem, as expected, is the lack of large annotated data. Table~\ref{tab:corpora}
shows all the POS-annotated CS corpora to our knowledge and their sizes. CS POS tagging
requires more annotated data compared to monolingual tagging, as CS increases 
the possible context of tokens.

\begin{table}[h!]

\begin{small}
\begin{tabular}{llrl}
Corpus		& Language     & Tokens       & Tag set \\
\hline \hline
S\&L'08	 & En-Es		& 8k 	& PTB\tablefootnote{\newcite{Solorio:2008b} report the tagset is a slightly modified version of PTB, but do not give the exact number of tags.} + 75 Es \\
V'14	& En-Hi			& 4k 	& 12 UT + 3 NE \\
J'15	& En-Hi			& 27k      & 34 Hi + 5 Twitter \\   
ICON'15\tablefootnote{Data from the ICON 2015 Shared Task on \textit{Pos Tagging For Code-mixed Indian Social Media Text}. It is available at \texttt{http://amitavadas.com/Code-Mixing.html}} & En-Hi			& 27k	& 34 Hi + 5 Twitter\\
	& En-Bn			& 38k	& 34 Hi + 5 Twitter\\
	& En-Ta			& 7k	& 17 UD \\
\c{C}\&\c{C}'16   & De-Tr	& 17k  	& 17 UD \\
S'16		& En-Hi		& 11k 	& 12 UT \\
S\&K'16		& midEn-La	& 3k 	& 12 UT
\end{tabular}
\caption{Overview of POS-annotated CS corpora. S\&L'08:\protect\newcite{Solorio:2008b}, V'14:\protect\newcite{vyas:2014},
J'15:\protect\newcite{jamatia:2015}, \c{C}\&\c{C}'16:\protect\newcite{cetinoglu:2016b},
S'16:\protect\newcite{sharma:2016}, S\&K'16:\protect\newcite{Schulz:2016b}.  UT:
Google Universal Tags \protect\cite{petrov:2012}. UD: Universal Dependencies 
tag set \protect\cite{nivre:2016}.}
\label{tab:corpora}
\end{small}
\end{table}

The last column of Table~\ref{tab:corpora} shows the tag sets used in annotating POS.
Only one corpus uses language-specific tags \cite{Solorio:2008b}, which predates
universal tag sets. With the introduction of Google Universal Tags (UT) \cite{petrov:2012}
and later, its extended version Universal Dependencies (UD) tag set \cite{nivre:2016}
preference has moved to using a common tag set for all tokens. \newcite{vyas:2014}
employ 3 additional tags for named entities. \newcite{jamatia:2015} and ICON 2015
Shared Task use a Hindi tag set that is mappable to UT. They 
also adopt 5 Twitter-specific tags.

\newcite{Solorio:2008b} show that  high accuracy 
English and Spanish taggers achieve only 54\% and 26\% accuracy respectively on their data,
indicating that off-the-shelf monolingual taggers are not suitable for CS text.
Common methods applied to overcome this problem in several experiments 
\cite{Solorio:2008b,vyas:2014,jamatia:2015,sharma:2016,Schulz:2016b}
are to choose between monolingual tagger outputs based on probabilities, 
utilising monolingual dictionaries and language models, and applying machine learning
on the annotated CS data.
One feature that deviates from standard POS tagging is language IDs, which are 
shown to be quite useful in previous work. Thus another challenge that comes with CS is 
predicting language IDs as a prior step to POS tagging. 

\newcite{Solorio:2008b} achieve a high score of 93.48\% with an SVM classifier, 
but this could be partly due to monolingual English sentences that constitute 
62.5\% of the corpus. In corpora with higher level of mixing, e.g. \cite{vyas:2014,jamatia:2015,sharma:2016}
best scores drop to 65.39\%, 72\%, and 68.25\% respectively. \newcite{Schulz:2016b} have 
an accuracy of 81.6\%. At the ICON 2015 Shared Task, the best system has an average
of 76.79\% accuracy. These scores show POS-tagging on CS data has room for improvement.

\section{Parsing}
\label{sec:parsing}
Parsing, the task of determining syntactic relations between words and phrases 
of a given sentence, has advanced substantially over the last decade. With the 
current rise of deep learning, a lot of parsers are developed, 
that, e.g. go above 93\% unlabelled attachment score in dependency parsing of English 
(cf. \newcite{Kiperwasser:2016} for a recent comparison of various high-performing
parsers).

While theories on parsing CS text have started quite early \cite{Joshi:1982} and 
a rule-based HPSG prototype is available \cite{goyal:2003}, there are no 
statistical parsers developed to handle CS. The main reason is the lack 
of treebanks that contain CS instances. Nevertheless, two recent works signal that
research is moving in this direction. 

\newcite{sharma:2016} build a pipeline for Hindi-English social media text. They 
create a corpus with four layers of annotation: language IDs, normalisation,
POS tags, and for the first time, chunk boundaries and labels. Each component of their pipeline
predicts one layer with data-driven approaches. When all steps are predicted
the accuracy for chunking is measured as 61.95\%.

\newcite{vilares:2016} train lexicalised bilingual parsers by merging the training
sets of two languages into one. They compare these models to monolingual ones 
on 10 Universal Dependencies treebanks \cite{nivre:2016}. 
The authors also apply their approach on English-Spanish code-switching
data in a toy experiment. They have annotated 10 tweets exhibiting CS according
to UD rules. They train an English-Spanish POS tagger by merging the training sets
of both languages. Their experiments show that using the bilingual tagger and 
parser is a promising direction for parsing CS.

Challenges in parsing CS stem from error propogation in previous steps, but 
also from the syntactic constructions that are not native to monolingual languages.
(\ref{ex:aufgabe}) is such an example from an ongoing corpus collection.

\begin{exe}
\begin{small}
\ex\label{ex:aufgabe} 
 \gll birka\c{c} \textit{Aufgabe}ler yapt{\i}k arkada\c{s}la \\
 {a few} assignment(DE).Pl(TR) make.Past.1Pl friend.Sg.Ins\\
 \trans `We made a few assignments with a friend.'  
\end{small} 
\end{exe}

The sentence above contains a German-Turkish mixed word \textit{Aufgabeler} (German
portion in italics) as 
explained in Section \ref{sec:langid} and the rest of the words are Turkish.
The whole sentence employs Turkish syntax, except that in the NP 
\textit{birka\c{c} Aufgabeler}, the noun modified by \textit{birka\c{c}} should be
 singular to be grammatical in Turkish. Perhaps the speaker utilises the 
German syntax where the noun is expected to be plural for this construction. 

(\ref{ex:wordorder}) is a similar example from \newcite{Broersma:2009} where 
the word order is from the embedded language (English, in italics),
and shows how the syntactic and lexical systems of the two languages 
are combined during production: lexical items of
one language are ordered according to the syntax of the other. 

\enote{sarah}{btw this totally contradicts the MLF theory that says that 
the syntax of an entire sentence follows the language of the inflexted verb.}
\enote{ozlem}{I didn't add it for the moment, because we do not mention Matrix
theory anywhere.}

\begin{exe}
\begin{small}
\ex\label{ex:wordorder}
    \gll \textit{Later} ik naaide voor mensen.\\
    Later I sewed for people.\\
\trans `Later I sewed for people.' \\
correct: `Later naaide ik voor mensen.'
\end{small}
\end{exe}

Although not explicitly CS, code-switching
bilinguals produce monolingual instances that do not follow the syntax of the 
uttered language. (\ref{ex:DEsyntax}) and (\ref{ex:NLsyntax}) show such instances 
where German syntax interferes with English \cite[p.130]{albrecht:2006} and 
Dutch syntax interferes with Turkish \cite{dogruoz:2009}. We include them into parsing challenges
as the CS corpora to be parsed is likely to contain similar monolingual constructions.

\begin{exe}
\begin{small}
\ex\label{ex:DEsyntax}
    Daniel: but me too not \\
    Faye: no, no, that goes not\\
    correct: `Daniel: but me neither \\
    Faye: no, no, that does not go'

\ex\label{ex:NLsyntax}
    \gll Beyimin ailesi hep o da burda.\\
     Husband.Gen family.Poss all it also here\\
\trans `My husband's family is also all here.' \\
correct: `Beyimin ailesi de hep burda.'
\end{small}
\end{exe}

Repeating a word or a whole clause in both languages in a loose or direct translation 
is a common CS phenomenon, especially in speech or historical documents, and
it might pose syntactic challenges e.g. when repetitive subordinate clauses 
exist which lead to complex coordinations  (\ref{ex:paternoster}) 
\cite[p.259]{Lodge:2008}. In this example the French portion (in italics) 
affirmatively translates and completes the Latin \textit{Pater Noster} verse.

\begin{exe}
\begin{small}
\ex\label{ex:paternoster}
    \gll Sed libera nos, \textit{mais} \textit{livre} \textit{nous,} \textit{Sire}, a malo, \textit{de} \textit{tout} \textit{mal} \textit{et} \textit{de} \textit{cruel} \textit{martire}\\
        But deliver us, but deliver us, God, from evil, of all evil and of cruel martyrdom\\   
    `But deliver us, God,  of all evil and martyrdom'
\end{small}    
\end{exe}

All these examples demonstrate that, in addition to the solutions that would 
improve preprocessing steps of parsing, new models and methods should be 
developed to handle parsing-specific problems. 

\section{Machine Translation}
\label{sec:mt}
Machine translation (MT) explores methods to translate text from one language to another.
Like all other tasks that rely on large amounts of data for training, MT quality decreases when encountering CS text. 
Not only the parallel text used for compiling phrase tables and translation probabilities but also the language models included are trained on monolingual data. 
Mixed text results in a high number of words unknown to the system and low probabilities for translations. 
Training dedicated translation systems for mixed text is, however, often not feasible due to the insufficient availability of data.

Moreover, translation quality increases with increasing context lengths (cf. phrase-based MT). 
CS, however, leads to limited accessibility of context (in form of phrases included as such in the phrase table) and thus leads to a decrease in translation quality.

A solution is to detect foreign words and then translate them into the matrix language before translating into a third language \cite{Sinha2005}. 
Identifying foreign material and translating them into the fitting word in context poses similar problems as described in Sections \ref{sec:normalisation} and \ref{sec:langid}. 
The lexical translations of inserted parts can be considered as a normalisation approach. 
In addition, an underlying assumption of this approach is the availability of a bilingual lexicon for the mixed language pairs which is not always a given.
Even in a perfect foreign word translation scenario, it is questionable if 
the ``monolingualised'' text syntactically and lexically behaves like any other monolingual text 
so that a conventional MT system can handle it.

Another challenge is intra-word CS due to morphological binding of one language to a stem from another language as often observed in e.g. Hindi-English text \cite{Sinha2005} or Turkish-German \cite{cetinoglu:2016a} which is shown in (\ref{ex:embtrans}) . 

\begin{exe}
\begin{small}
\ex\label{ex:embtrans} 
 \gll Lehrerzimmer\textit{de} schokolade \textit {da\u{g}ıtıyorlar}\\
      {staff room}(DE).Loc(TR) chocolate {distribute.Prog.3Pl}\\
   `They give away chocolate in the staff room.'

\ex\label{ex:turkishtrans} 
Handing schokolade Lehrerzimmer
\end{small}
\end{exe}

Google translate\footnote{ Google Translate, \texttt{translate.google.com}, 29.07.16.}
returns (\ref{ex:turkishtrans}) as a translation from Turkish to English. The Turkish morpheme \textit{de} 
is correctly recognised as an inflectional suffix and severed from the base word \textit{Lehrerzimmer} `staff room'.
Yet, it is not translated as the preposition \textit{in} as expected. The German word is present, but without a translation.
Moreover, the subject of the sentence (which should be \textit{they})
 is not translated at all even though the information is contained in the purely 
 Turkish word \textit{da\u{g}ıtıyorlar} `they distribute'. Another word oblivious 
 to the translation is \textit{schokolade} `chocolate'. When the same sentence
 is input to Google Translate, all in German and all in Turkish, both cases receive
 a fully-translated output.\footnote{The outputs are grammatical in both languages
 yet the semantics do not exactly match the original sentence.}
 Thus, the mixed context seems to harm the correct translation of the sentence.


\newcite{Manandise:2011} describe a way to deal with morphological bindings in the context of MT. They use a morphological analyser to first separate the base 
word from its morphological affixes. Those are then analysed and translated according to the morphology of the target base. They give Example (\ref{ex:borrow}), an English-Spanish mixed word \textit{anticooldad} `anti-coolness':

 \begin{exe}
\begin{small}
\ex\label{ex:borrow} 
\begin{xlist}
      \ex\label{ex:borrow1} anticooldad\\
      \ex\label{ex:borrow2} anticool: dad\\
      \ex\label{ex:borrow3} cooldad:anti\\
      \ex\label{ex:borrow4} cool: anti, dad\\
      \ex\label{ex:borrow5} dummy: anti, dad\
      
       \end{xlist}
 
\end{small}
\end{exe}

The analyser returns all possible base terms (shown as the string before the colon in (\ref{ex:borrow})) along with the 
possible morphemes (shown as the strings after the colon in (\ref{ex:borrow})). Since the 
word \textit{cool} appears in the English dictionary and the other suggested base terms do not, 
the translation of the morphemes along with the correct morphological analysis and language-specific rules lead to the translation \textit{anti-coolness}.

Even though there might be suggested solutions for token-based translations of embedded words as a preprocessing step,
translation of the monolingualised sentence might still pose problems due to syntactic specificities as described in Section~\ref{sec:parsing}. In case the sentence is monolingualised into one language and uses the syntax of the other original language, MT faces the problem of two separate combined systems: the lexical system of one language and the syntax of another.

\section{Automatic Speech Recognition}
\label{sec:asr}
For automated processing of spoken communication, an automatic speech recognition (ASR) system, which transforms speech signal to text, is an essential component. 
In the context of CS, ASR is important because CS appears mainly in conversational speech.
To develop an ASR system, three major components need to be built: a pronunciation dictionary, a language model and an acoustic model (AM) \cite{Young1996}.
In general, there are two possible ways to build an ASR system for CS speech \cite{Vu2012}. 
In the first approach, a LID system is used to split the CS speech into monolingual parts and, afterwards, monolingual recognisers are applied to the corresponding speech segments. 
This method is straightforward since the monolingual systems are already available. 
We lose, however, the semantic information between the segments and the mistakes of the LID system cannot be recovered especially if speech segments are short (e.g. < 3s). 
The second approach is building an integrated system with a multilingual AM, dictionary and language model. 
Compared to the first approach, it allows handling of CS within a word and the semantic information can be used between languages. 
Therefore, we focus only on identifying challenges of developing pronunciation dictionaries and acoustic models for multilingual ASR.

\noindent
\textbf{Pronunciation dictionary} A pronunciation dictionary is a collection of words and phoneme sequences which describe how a word is pronounced.
A straightforward implementation of a dictionary is to combine pronunciations of the mixed languages.
This is often not suitable because pronunciation often changes in a CS context due to the articulation effect when speakers switch from one language to another.
Another challenge is how to automatically create the pronunciation for CS words. 
To our best knowledge, this is a difficult task which has never been addressed so far.

\noindent
\textbf{Acoustic modelling} An AM estimates the probability of sound state given a speech frame.
The most crucial problem is again the lack of transcribed CS data.
Another one is the phonetic transfer phenomenon which occurs even when the speaker is proficient in both languages.
Hence, most recent proposed approaches focus on bilingual acoustic models which combine the properties of both languages and to some extent overcome the sparsity problem.
\newcite{Vu2012} merge the phoneme sets based on the International Phonetic Alphabet (IPA) manual mapping to share training data between phonemes across languages.
Furthermore, their system allows to ask language specific questions during the context dependent decision tree building process.
They achieve an improvement over the baseline with concatenated phoneme sets.
\newcite{Li2013} propose an asymmetric AM which automatically derives phone clusters based on a phonetic confusion matrix.
In the decision tree building process, they identify similar context dependent tree states across languages. 
The new output distribution is a linear interpolation of the pretrained monolingual state models.
Their proposed approach outperforms the baseline with a large margin.

Another direction is to integrate LID prediction into ASR during testing. 
The LID gives the probability of a language given a speech frame which can be combined directly with the acoustic score for testing. 
\newcite{Weiner2012} show good improvement when the LID system is sufficiently accurate.
It is, however, a challenging task to develop a LID system on the acoustic frame level.

\section{Conclusion}
\label{sec:conclusion}

In this paper, we discuss the challenges that surface when well-established NLP
tasks deal with CS data. Some of these challenges are language-pair dependent e.g.
Romanisation and back-translation in Hindi or Bengali. Others are recurring
throughout various tasks regardless of the language such as the increased amount 
of unseen constructions caused by combining lexicon and syntax of two languages.
 
Working on NLP for mixed data yields the advantage that resources and tools 
for the respective languages can be beneficial. Although we do not have to 
start from scratch, the tasks and required techniques are significantly different 
from those for monolingual data.
Context-sensitive methods suffer due to increased combinatoric possibilities crossing syntactic and lexical systems of 
different languages.

In addition, CS is a phenomenon that appears in data with hard-to-process factors other than mixing.
CS-typical genres are often close to spoken text and thus have to deal 
with problems that colloquial text poses from non-canonicity to incomplete syntactic structures to 
OOV-words. Although this would already suggest that a higher number of training instances are needed,  there 
are just small amounts of annotated data available.
So far there are annotated bilingual training resources for just three of the tasks (LID, POS and ASR)
for specific language pairs. Since each mixed language comes with its own challenges, each pair requires a dedicated corpus.

To alleviate the data sparsity problem, some approaches work by generating artificial CS text based on a CS-aware recurrent neural network decoder \cite{Vu2014} or a machine translation system to create CS data from monolingual data \cite{Vu2012}. 
Such techniques would benefit from better understanding of the characteristics of code-switching data. This is why 
we enriched our paper with examples from data sets covering different language pairs.
So far, very little NLP research makes use of linguistic insights into CS patterns (cf. \newcite{Li2014}). 
Such cues might improve results in the discussed tasks herein.

Another recurring and not yet addressed issue\footnote{\newcite{vyas:2014}
discuss joint modelling in the conclusion.}, is the inter-relatedness of all these tasks.
Features required for one task are the output of the other. Pipeline approaches 
cannot take advantage of these features when task dependencies are cyclic (e.g., normalisation and language identification). 
Moreover pipelines cause error propagation. 
This fact asks for attention on joint modelling approaches.

\section*{Acknowledgments}
We thank our anonymous reviewers for their helpful comments. This work was funded 
by the Deutsche Forschungsgemeinschaft (DFG) via SFB 732, projects D2 and A8, and 
by the German Federal Ministry of Education and Research (BMBF) via CRETA Project.

\bibliography{csbiblio}
\bibliographystyle{emnlp2016}

\end{document}